%% file: arxiv.tex
\documentclass[10pt,twocolumn,letterpaper]{article}

\usepackage[pagenumbers]{cvpr} 

\usepackage{graphicx}
\usepackage{amsmath}
\usepackage{amssymb}
\usepackage{booktabs}
\usepackage{enumitem}
\usepackage{multicol}
\usepackage{multirow}

\usepackage{cinzel}
\usepackage[T1]{fontenc} 
\newcommand{\modelname}{\textcinzel{\textit{MIST }}}
\newcommand{\textcinzelit}[1]{\textcinzel{\textit{#1}}}
\newcommand{\supplement}{appendix }

\usepackage[pagebackref,breaklinks,colorlinks]{hyperref}

\usepackage[capitalize]{cleveref}
\crefname{section}{Sec.}{Secs.}
\Crefname{section}{Section}{Sections}
\Crefname{table}{Table}{Tables}
\crefname{table}{Tab.}{Tabs.}

\def\cvprPaperID{3270} 
\def\confName{CVPR}
\def\confYear{2023}

\begin{document}

\title{\modelname: Multi-modal Iterative Spatial-Temporal Transformer for Long-form Video Question Answering}

\author{
    Difei Gao\textsuperscript{\rm 1},
    Luowei Zhou\textsuperscript{\rm 2}\thanks{Currently at Google Brain.} , 
    Lei Ji\textsuperscript{\rm 3},
    Linchao Zhu\textsuperscript{\rm 4},
    Yi Yang\textsuperscript{\rm 4}, 
    Mike Zheng Shou\textsuperscript{\rm 1}\thanks{Corresponding author.} \\
    \\
        \textsuperscript{\rm 1}Show Lab, National University of Singapore, 
    \textsuperscript{\rm 2}Microsoft, \\
    \textsuperscript{\rm 3}Microsoft Research Asia,
    \textsuperscript{\rm 4}Zhejiang University
}
\maketitle

\begin{abstract}
To build Video Question Answering (VideoQA) systems capable of assisting humans in daily activities, seeking answers from long-form videos with diverse and complex events is a must. Existing multi-modal VQA models achieve promising performance on images or short video clips, especially with the recent success of large-scale multi-modal pre-training. However, when extending these methods to long-form videos, new challenges arise. On the one hand, using a dense video sampling strategy is computationally prohibitive. On the other hand, methods relying on sparse sampling struggle in scenarios where multi-event and multi-granularity visual reasoning are required. In this work, we introduce a new model named \textcinzelit{M}ulti-modal\textcinzelit{I}terative\textcinzelit{S}patial-temporal\textcinzelit{T}ransformer\textcinzel{\textit{(MIST})} to better adapt pre-trained models for long-form VideoQA. Specifically, \modelname decomposes traditional dense spatial-temporal self-attention into cascaded segment and region selection modules that adaptively select frames and image regions that are closely relevant to the question itself. Visual concepts at different granularities are then processed efficiently through an attention module. In addition, \modelname iteratively conducts selection and attention over multiple layers to support reasoning over multiple events. The experimental results on four VideoQA datasets, including AGQA, NExT-QA, STAR, and Env-QA, show that \modelname achieves state-of-the-art performance and is superior at computation efficiency and interpretability.

\end{abstract}

\input{section/intro}

\input{section/rel}

\input{section/method}

\input{section/exp}

\input{section/conclusion}

\renewcommand\thesection{\Alph{section}}
\setcounter{section}{0}

\section*{Appendix}
In the appendix, we provide additional details for the main paper:
\begin{itemize}[itemsep=2pt,topsep=3pt,parsep=0pt]
    \item More discussions of top-k selector in our proposed model, \modelname, in Sec.~\ref{secA}.
    \item More details of experimental settings in Sec.~\ref{secB}.
    \item More experimental results in Sec.~\ref{secC}.
    \item More visualizations of prediction results in Sec.~\ref{secD}.
\end{itemize}

\section{More Discussions of Top-k Selector}
\label{secA}
As illustrated in the main paper, \modelname calculates multi-modal attention between segment/patch features and question features, then performs top-k hard selection over segment/patch features. Commonly used hard selection, such as argmax or top-k selection functions, will stop the back-propagation of gradients and affect the training of the attention module. Thus, we use the Gumbel-Softmax trick to perform the differentiable hard selection. Note that the standard Gumbel-Softmax trick is designed for top-1 selection, and extending it to top-k selection is still an open problem. 

The core of extending Gumbel-softmax to top-k selection is multiple sampling. The main difference between different implementations is whether sample with replacement. Many previous works~\cite{kool2019stochastic, kool2020ancestral, struminsky2021leveraging} for sequence generation tasks choose to sample without replacement to generate diverse sequences, i.e., sampling the first element, then renormalizing the remaining probabilities to sample the next element, etcetera. But as also mentioned in their papers, sampling without replacement means that the inclusion probability of element $i$ is not proportional to $p_i$. For an extreme example, if we sample $k = n$ elements, all elements are included with probability 1. It may affect their word sequence generation tasks a little, because the number of candidate words in each step is usually much larger than top-k, so all top-k choices could be plausible. However, for our targeted long-form VideoQA task, the model needs only to select the segment related to the question. And in some cases, the question could only involve one segment. Thus, we expect the model learns to enhance the most related segment in such cases by re-sampling it, instead of forcing it to select an irrelevant segment.

\section{More Details about Experiment Setting}
\label{secB}
\textbf{Evaluation Metrics.}
In the main paper, we evaluate our method on AGQA v2~\cite{grunde2022agqa}, NExT-QA~\cite{xiao2021next}, STAR~\cite{wu2021star}, and Env-QA~\cite{gao2021env}. In the supplement, we additionally evaluate \modelname on AGQA v1~\cite{grunde2021agqa}. For the AGQA v1/v2, NExT-QA, and STAR, we follow their paper using the QA accuracy as the metric, i.e., if the answer is the same as the ground truth, then the model gets 1 score; otherwise gets 0. For Env-QA, we follow the dataset paper that first decomposes the phrase answer into several parts, called role-value format, then uses an IOU-like score to evaluate the similarity between the prediction and ground truth.

\textbf{Implementation Details.}
As illustrated in Sec. 3.3 of the main paper, we calculate the similarity between the answer candidates and the combining feature of video and question to predict the answer. For AGQA v1/v2, NExT-QA, and STAR, the answer prediction follows the above-mentioned method. But for Env-QA, since the model needs to predict a set of role-value answers, we slightly modify the answer prediction module, i.e., plug a set of classifiers to predict the answer in each role. Then, we sum the losses of all classifiers to train the whole model.

\section{Experimental Results}
\label{secC}
\textbf{Performances on AGQA v1.}
Since the dataset creators~\cite{grunde2021agqa} of AGQA recommend using its v2 version, we evaluated our method and conducted ablation studies on AGQA v2 in the main paper. Here, we report the results on AGQA v1 to compare with some recent works~\cite{xiao2022video, qian2022dynamic, lee2022dense} which didn't report their results on AGQA v2. 

From the results in Table~\ref{tab:agqa_v1}, we can see that \modelname outperforms state-of-the-art methods by a large margin. Compared to the previous SOTA, which uses the same feature, i.e., Temp[ATP], \modelname obtains about 4\% performance boost. In addition, ~\cite{qian2022dynamic, lee2022dense} proposed specific designs, such as modular networks and dependency attention modules, tailored to compositional questions in AGQA. \modelname achieves better compositional reasoning ability with a more general design for long-form video QA.

\begin{table}[]
\centering
\resizebox{0.355\textwidth}{!}{%
\begin{tabular}{l|cc|c}
\toprule
\multicolumn{1}{c|}{Method}    & Binary & Open  & All      \\ \midrule
PSAC~\cite{li2019beyond}       & 54.19 & 27.20 & 40.40                \\
HCRN~\cite{le2020hierarchical}        & 58.11 & 37.18 & 47.42       \\
HME~\cite{fan2019heterogeneous}       & 59.77 & 36.23 & 47.74                \\
DualVGR~\cite{wang2021dualvgr}        & 55.48 & 40.75 & 47.80       \\
HQGA~\cite{xiao2022video}             & 56.15 & 39.49 & 47.48       \\
DSTN-E2E~\cite{qian2022dynamic}       & 57.38 & 42.43 & 49.60       \\
Temp[ATP]~\cite{buch2022revisiting}   & 59.60  & 49.16 & 54.17      \\ \midrule
\modelname- CLIP                      & 63.36 & 53.51 & 58.26    \\ \bottomrule
\end{tabular}%
}
\vspace{-0.3cm}
\caption{QA accuracies of state-of-the-art methods on AGQA v1 test set.}
\label{tab:agqa_v1}
\vspace{-0.3cm}
\end{table}

\textbf{Comparison of Different Top-k Selectors.}
We try different top-k selectors to show the effect of different implementations:
\begin{itemize}[itemsep=2pt,topsep=2pt,parsep=0pt,leftmargin=12pt]
    \item G-S w. Replacement: We sample the region features of $Top_k$ segments by sampling segment indexes with replacement $Top_k$ times. It is our default setting in the main paper.
    \item G-S w/o. Replacement: We sample the region features of $Top_k$ segments by sampling segment indexes without replacement $Top_k$ times. Specifically, after each selection, the probabilities of selected segments on next round sampling are set to 0.
    \item Non-params. Attention: Another naive solution for addressing training issue of attention module is to propose a non-parametric attention module. Specifically, the selector decides whether a certain segment should be selected based on the matching score of its pre-trained visual feature and the given question feature. The pre-trained feature already has a certain matching ability. Thus, we first normalize the pre-trained visual features and question features, then perform a dot product to obtain the attention score. 
    Note that there is one limitation for this model, because the attention strategy cannot be trained, so no matter how many iterations, the selected segments are all the same. So, this variant has only one layer of ISTA.
\end{itemize}

\begin{table}[]
\centering
\resizebox{0.475\textwidth}{!}{%
\begin{tabular}{l|c|ccc}
\toprule
\multicolumn{1}{c|}{Method}    & \# of ISTA layer & Binary & Open  & All      \\ \midrule  
G-S w/o. Replacement  & 2 & 58.28 & 50.56     &  54.39  \\ 
G-S w. Replacement   & 2 & 57.68 & 50.06  & 53.84   \\ 
G-S w. Replacement   & 1 & 56.84 & 49.18  &52.98  \\ 
Non-params. Attention  & 1 & 56.34  & 49.52  & 52.91   \\ 
\bottomrule 
\end{tabular}%
}
\vspace{-0.3cm}
\caption{QA accuracies of \modelname with different top-k selectors on AGQA v2 test set.}
\label{tab:selector}
\vspace{-0.3cm}
\end{table}

From the results in Table~\ref{tab:selector}, we can see that sampling with replacement slightly outperforms the one without replacement. This may be because of the issue we mentioned in Sec~\ref{secA}, i.e., sampling without replacement may introduce some irrelevant segments that disturb the reasoning. In addition, we can see that Non-params. Attention achieves similar performance with G-S w. Replacement with 1 ISTA layer, but is worse than G-S w. Replacement with 2 ISTA layers. It indicates that pre-trained features do already have a certain matching ability, but we need to modify it to support iterative attention. 
    
\begin{figure*}[t]
\centering
\includegraphics[width=0.87\textwidth]{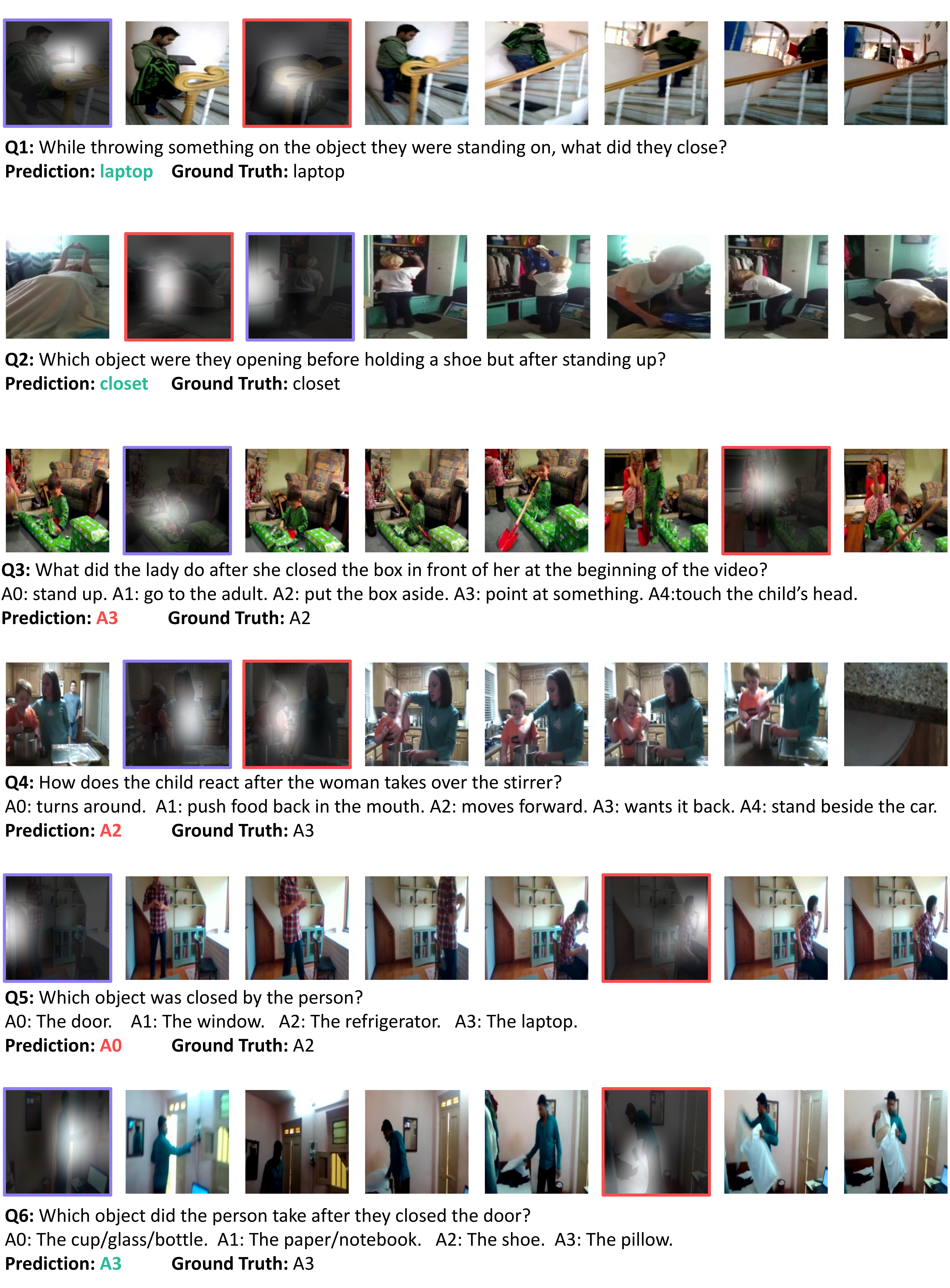}
\caption{\textbf{Qualitative results of \modelname.} We visualize its prediction results along with spatial-temporal attention, where the frames with purple and red outlines indicate the highest temporal attention score in the first and second ISTA layers, respectively.}
\label{fig-sup.example}
\end{figure*}

\section{More Visualizations of MIST}
\label{secD}

In Figure~\ref{fig-sup.example}, we show more visualizations of the prediction results of \modelname. It can be seen that our model can select video clips and image regions relevant to the question. We also find the model to be wrong in the following cases: 1) The object is partially visible, and the model needs to infer what the object is from the video context. For example, in Q3, a partially visible lady puts the box aside at the beginning of the video. The lady is fully visible only in the latter frames. So, given a question requiring locating segments where a lady appears in the beginning, the model incorrectly finds the segment with a fully visible lady. 
2) The video contains a large number of similar events. It is hard for our model to distinguish these subtle differences. In Q5, the boy first stirs the food in the pot, then the lady takes over the stirrer, and the boy pulls the lady's arm to want to get the stirrer back. All these events are quite similar, and it is still relatively hard to locate the correct segments with the given question. 

{\small
\bibliographystyle{ieee_fullname}
\bibliography{egbib}
}

\end{document}

%% file: section/intro.tex
\section{Introduction}
\label{sec:intro}

One of the ultimate goals of Video Question Answering (VideoQA) systems is to assist people in solving problems in everyday life~\cite{grauman2022ego4d, wong2022assistq, lei2021assistsr}, e.g., helping users find something, reminding them what they did, and assisting them while accomplishing complex tasks, etc. To achieve such functions, the systems should be able to understand and seek the answer from long-form videos with diverse events about users' activities.

\begin{figure}[t]
\includegraphics[width=0.475\textwidth]{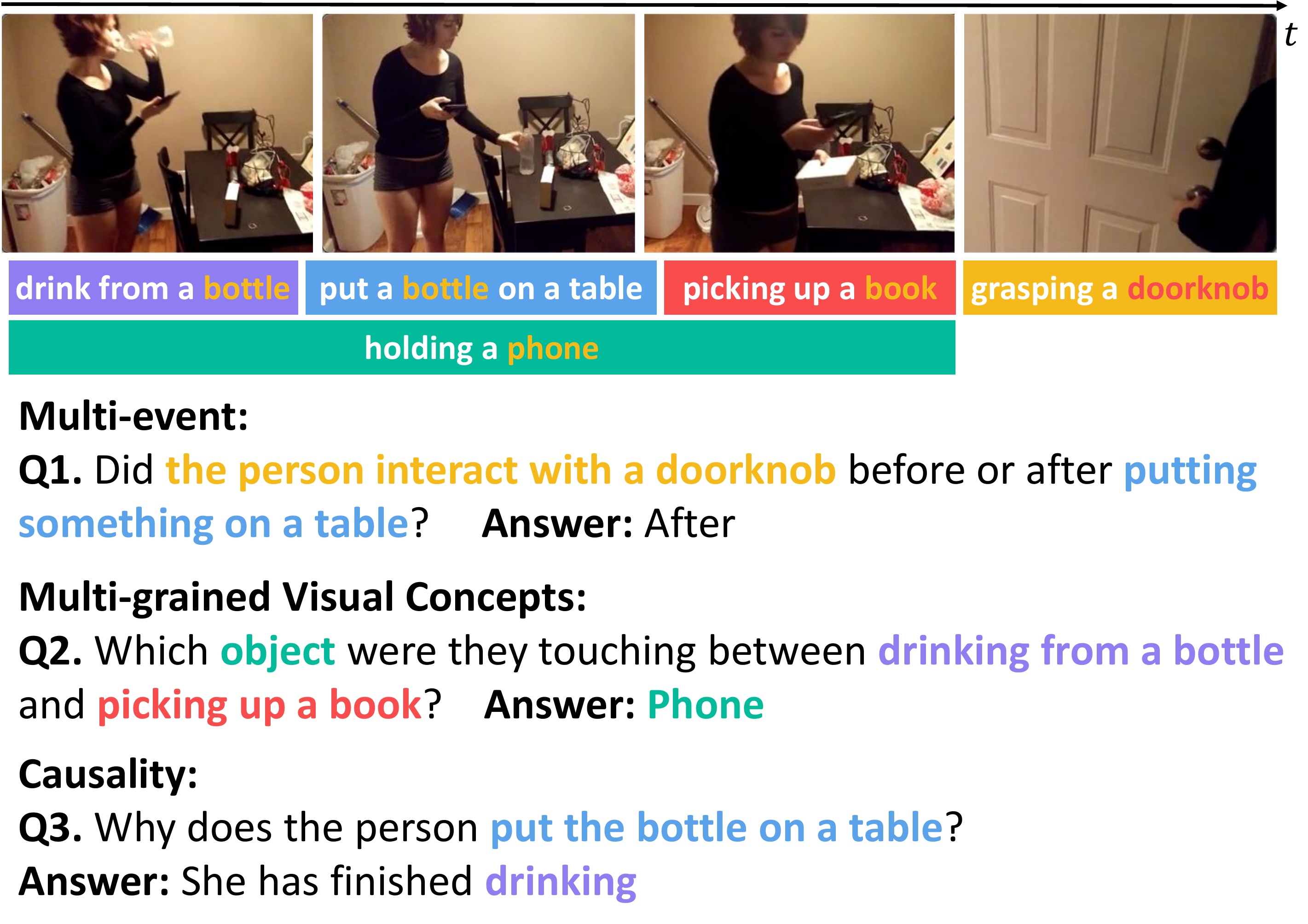}
\vspace{-0.5cm}
\caption{\textbf{Main challenges of long-form VideoQA.} The questions for long-form VideoQA usually involve multi-event, multi-grained, and causality reasoning.}
\vspace{-0.3cm}
\label{fig.teaser}

\end{figure}

Compared to understanding and reasoning over short videos, many unique challenges arise when the duration of the video increases, as shown in Fig.~\ref{fig.teaser}: 1) Multi-event reasoning. The long-form videos usually record much more events. The questions about these videos thus naturally require the systems to perform complex temporal reasoning, e.g., multi-event reasoning (Q1 in Fig.~\ref{fig.teaser}), causality (Q3), etc. 2) Interactions among different granularities of visual concepts. The questions of short-clip videos usually involve the interactions of objects or actions that happened simultaneously, while questions for long-form videos could involve more complex interactions of objects, relations, and events across different events, e.g., Q2 in Fig.~\ref{fig.teaser}.

\begin{figure}[t]
\centering
\includegraphics[width=0.47\textwidth]{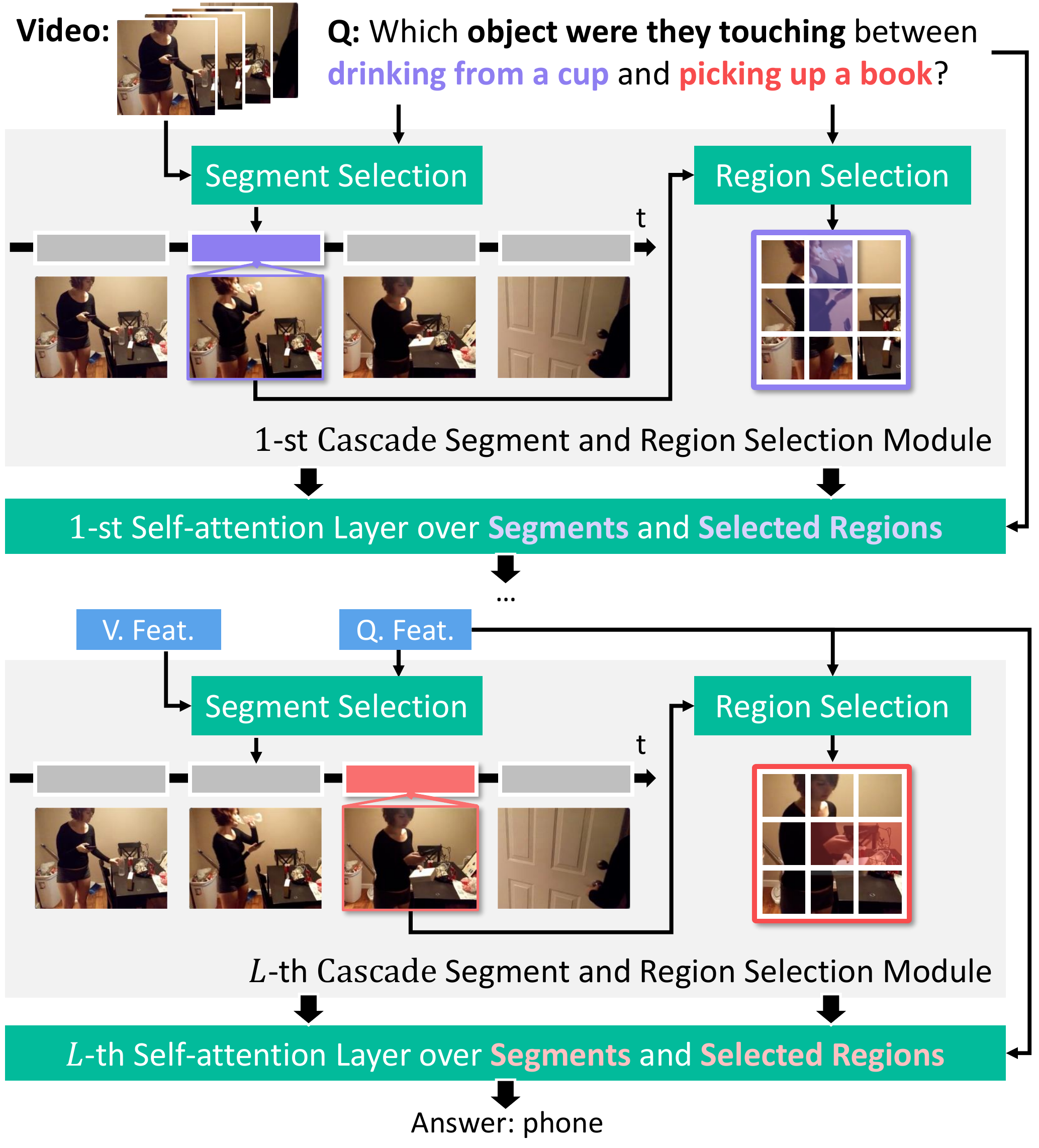}
\vspace{-0.3cm}
\caption{\textbf{Diagrammatic illustration of\modelname.} It revises a standard spatial-temporal self-attention layer into two modules: a cascade selection module that dynamically eliminates question-irrelevant image regions, and a self-attention layer reasoning over multi-modal multi-grained visual concepts. The proposed modules further iterate multiple times to reason over different events.}
\label{fig:2}
\vspace{-0.3cm}
\end{figure}

Current vision-language methods~\cite{lu202012, chen2020uniter, li2020oscar, lei2021less, fu2021violet, bain2021frozen, luo2022clip4clip, zellers2021merlot, zellers2022merlot} excel at QA over images or short clips spanning several seconds. In other words, they excel at learning multi-modal correspondences between a single caption with one or few events. Their tremendous progress over these years is fueled by 1) pre-training on large-scale image-language~\cite{sharma2018conceptual, krishna2017visual, radford2021learning} and short-clip-language datasets~\cite{miech2019howto100m, bain2021frozen}, and 2) end-to-end multi-modal Transformers~\cite{radford2021learning, bao2021beit, fu2021violet, bain2021frozen, wang2022all, alayrac2022flamingo}, which is superior at learning the alignments between image patches with texts. 

However, these multi-modal Transformers rely on dense self-attention, with the computation cost increasing exponentially over time, especially when adapting to long-form videos. To make the dense self-attention computationally feasible in processing videos, almost all current state-of-the-art pre-trained Transformers are sparse sample-based methods, e.g., ~\cite{bain2021frozen, wang2022all} only sample 3 or 4 frames per video regardless of its length. If we simply adapt these pre-trained models to long-form videos with the same sampling strategy, there will be a domain gap between the pre-training and downstream VideoQA tasks. In pre-training, the sparsely sampled frames of a short video depict a coherent action, while they are likely to be random shots for part of events in a long video. Recently, some early attempts process the video hierarchically~\cite{buch2022revisiting}, which splits the video into several segments and performs QA only on aggregated segment-level features. It can ease the efficiency issue, but it is still hard to capture complex interactions among multi-grained concepts. Therefore, leveraging the advantages of models pre-trained from images and short videos and, in the meantime, addressing the unique challenges of long-form VideoQA is worth exploring.

In this paper, we propose a new model, named\textcinzelit{M}ulti-modal\textcinzelit{I}terative\textcinzelit{S}patial-temporal\textcinzelit{T}ransformer\textcinzel{\textit{(MIST})}, as shown in Fig.~\ref{fig:2}.\modelname comes from a simple finding that for long-form VideoQA, it is not necessary to consider the details of all events in a video, like what dense self-attention over all patches do. The model only needs to consider the general content of all events and focuses on the details of a few question-related events. Thus, \modelname decomposes dense joint spatial-temporal self-attention into a question-conditioned cascade segment and region selection module along with a spatial-temporal self-attention over multi-modal multi-grained features. The cascade selection reduces the computation cost and benefits the performance by focusing on the question-related segments and regions. The self-attention over segments and image patches, better captures interactions among different granularities of visual concepts. In addition, through iteratively conducting selection and self-attention,\modelname can reason over multiple events and better perform temporal and causal reasoning.

We conduct experiments on several recently released VideoQA datasets with relatively longer videos, AGQA~\cite{grunde2021agqa},  NExT-QA~\cite{xiao2021next}, STAR~\cite{wu2021star}, and Env-QA~\cite{gao2021env}, with an average video duration varies from 12s to 44s. The experimental results show that our approach achieves state-of-the-art performance on all four datasets. Further ablation studies verify the effectiveness of the key components in our approach. Moreover, quantitative and qualitative results also show that our method provides higher efficiency and interpretability, providing reasonable evidence for answering questions.

%% file: section/rel.tex
\section{Related Work}
\label{sec:relate}

\textbf{Video question answering.} 
Video Question Answering is one typical type of vision-language task studied for many years. Some datasets~\cite{jang2017tgif, xu2017video} focus on short clips about daily human activities, e.g., sports, household work, etc. Some others, such as TVQA~\cite{lei2018tvqa}, MovieQA~\cite{tapaswi2016movieqa}, and Social-IQ~\cite{zadeh2019social}, mainly focus on long videos cropped from movies or TV series for evaluating the understanding of the plot and social interactions, where subtitles play an essential role. Recently, ~\cite{grunde2021agqa, xiao2021next, wu2021star, gao2021env} aim to evaluate more complex spatial-temporal reasoning ability over long-form videos, e.g., causality, sequential order, etc. Current works achieve promising results on the first two types of benchmarks, while they still struggle on the last one, which is the main focus of this paper.

In terms of methodology, early-stage works proposed various LSTM or Graph Neural Network-based models to capture cross-modal~\cite{li2019beyond, park2021bridge, zhao2018open} or motion-appearance interaction~\cite{le2020hierarchical, gao2018motion}. One recent work~\cite{xiao2022video} integrates graph modeling into Transformers to explicitly capture the objects and their relations in videos. In addition, with the great success of pre-trained vision-language Transformers, many works~\cite{bain2021frozen, fu2021violet, wang2022all} directly fine-tune the pre-trained model on downstream VideoQA tasks. ~\cite{buch2022revisiting} proposes a simple yet effective fine-tuning strategy to hierarchically process videos with pre-trained Transformers.

Compared to previous works, this paper is an early attempt to specifically focus on the challenges of long-form VideoQA for Transformers-based methods. Specifically, we revise the self-attention mechanism to better perform multi-event, multi-grained visual concepts reasoning. 

\textbf{Transferring pre-trained models to downstream tasks.}
Many works try to transfer pre-trained vision-language Transformers, such as CLIP~\cite{radford2021learning}, into downstream tasks, e.g., object detection~\cite{gu2021open}, image generation~\cite{patashnik2021styleclip}, and video-text retrieval~\cite{luo2022clip4clip, xue2022clip, zhao2022centerclip, fang2021clip2video}. CLIP4Clip~\cite{luo2022clip4clip} proposes various aggregation methods for CLIP features, e.g., mean pooling, Transformer, to better represent a video. CLIP2Video~\cite{fang2021clip2video} proposes a temporal difference block to better capture motion information.

Similar to the above methods, we preserve the strengths of pre-trained models and improve their weaknesses on downstream tasks, but this works focus on another one, long-form VideoQA, where the main focus is on multi-event and multi-granularity reasoning.

\textbf{Long-form video modeling.}
With the great success of short-term video understanding in recent years, some pioneer works~\cite{wu2021towards, cheng2022tallformer} have started to focus on long-form video modeling for action recognition or localization tasks. They mainly focus on increasing the efficiency of processing long video features. ~\cite{cheng2022tallformer} proposes short-term feature extraction and long-term memory mechanisms that can eliminate the need for processing redundant video frames during training. ~\cite{lin2022eclipse} proposes to replace parts of the video with compact audio cues to succinctly summarize dynamic audio events and are cheap to process. ~\cite{islam2022long} introduces structured multi-scale temporal decoder for self-attention to improve efficiency.

The above methods utilize the natural characteristics of videos to reduce the computation. In contrast, this paper considers the characteristics of QA tasks, using the question as a guide to reduce computation and benefit performance.

\textbf{Iterative Attention.}
Similar to our method, there are also some existing works~\cite{mnih2014recurrent, jaegle2021perceiver, bertasius2021space} for designing iterative attention mechanisms to increase the computation efficiency or obtain a better performance. ~\cite{mnih2014recurrent} proposes a recurrent image classification model to iteratively attending on a sequence of regions at high resolution.  Perceiver~\cite{jaegle2021perceiver} revises self-attention in Transformer to an asymmetric attention to iteratively distill inputs into a tight feature, allowing it to handle large inputs. TimeSformer~\cite{bertasius2021space} proposes various self-attention schemes for video classification models to separately apply temporal and spatial attention. Our model differs in that it utilizes multi-modal correspondence (i.e., both visual signal and question) in VideoQA to guide iterative attention.

%% file: section/method.tex
\section{Method}

\begin{figure*}[t]
\includegraphics[width=\textwidth]{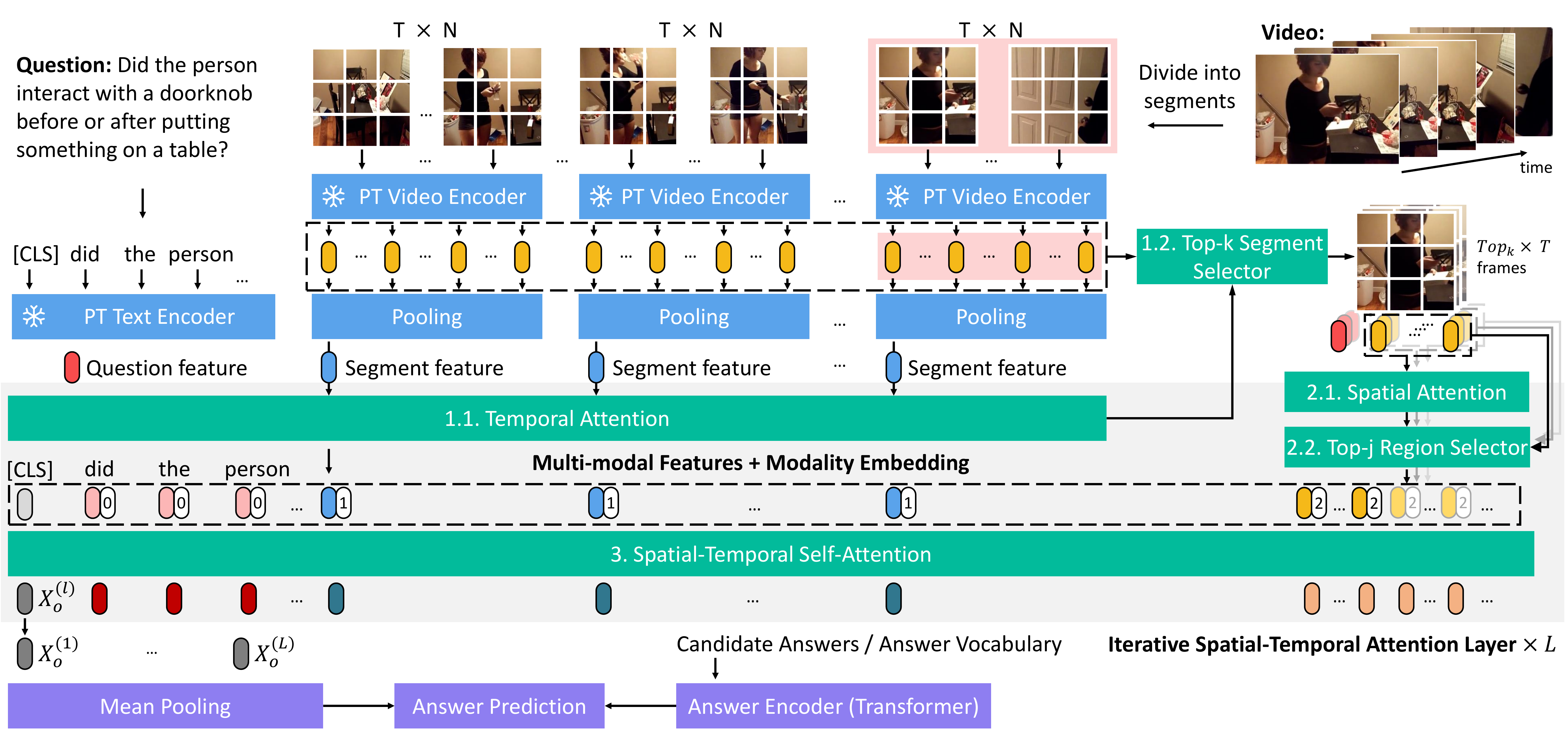}
\vspace{-0.7cm}
\caption{\textbf{Architecture of \modelname.} \modelname first divides video into several segments and utilizes the pre-trained model to extract the feature of each one. Then,\modelname iteratively performs self-attention over a selected set of features to reason over multiple events. Finally, it predicts the answer by comparing the combination of video and question features with answer candidate features. Note that the "PT Video Encoder" in the figure can also be image-based encoders.}
\label{fig:model}
\vspace{-0.3cm}
\end{figure*}

The goal of a VideoQA task is to predict the answer $y$ for a given video $\mathcal{V}$ and a question $q$, formulated as follows:
\begin{align}
    \widetilde{y} = { \underset{y\in \mathcal{A}}{\arg \max}} {\mathcal{F}_{\theta}(y|q, \mathcal{V}, \mathcal{A}}),
\end{align}
where $\widetilde{y}$ is the predicted answer chosen from the candidate answers (i.e., answer vocabulary or provides choices), denoted as $\mathcal{A}$, and $\theta$ is the set of trainable parameters of a VideoQA model $\mathcal{F}$.

In Fig.~\ref{fig:model}, we present the pipeline of our proposed \textcinzelit{M}ulti-Modal\textcinzelit{I}terative\textcinzelit{S}patial-temporal\textcinzelit{T}ransformer,\textcinzel{\textit{MIST}}.\modelname answers the question in three steps: 1) utilize a pre-trained model to extract the input features, 2) iteratively perform self-attention over a selected set of features to perform multi-event reasoning, 3) predict the answer based on the obtained video, question, and answer features.

\subsection{Input Representation}
\label{sec3.1}
Existing vision-language Transformers are good at representing images or short clips. To adapt them to handle the long-form video, we first split the video into $K$ uniform length segments, where each segment contains $T$ frames. In addition, each frame is divided into $N$ patches. Note that, for the simplicity of notation, the [CLS] token for image patches and frames are counted in $N$ and $T$.

The vision-language Transformer, like CLIP, All-in-one, with frozen parameters, extracts patch-level features of all segments, $\boldsymbol{x} = \{x^1, x^2, ..., x^K\}$,  where $x^k \in \mathbb{R}^{T \times N \times D}$ is the feature of $k$-th segment, where $D$ is the dimension of each patch-level feature. The patch-level visual token features will be used to obtain frame and segment features in the following modules. Since the segment features are separately extracted, to indicate their temporal positions in the whole video, we add position embedding $P_t \in \{\phi_t(i) | i \in [0, K \cdot T]\}$ for each token with their frame index.

For the text part, the question is tokenized as a sequence of words, and then fed into the vision-language Transformer to get word-level features $\mathbf{X_w} = \{w_1, ..., w_M\}$, where $w_1$ corresponds to $[\mathrm{CLS}]$ and $w_2, ..., w_M$ are words in question.

\subsection{Iterative Spatial-Temporal Attention Layer}
\label{sec3.2}

\begin{figure}[t]
\centering
\includegraphics[width=0.475\textwidth]{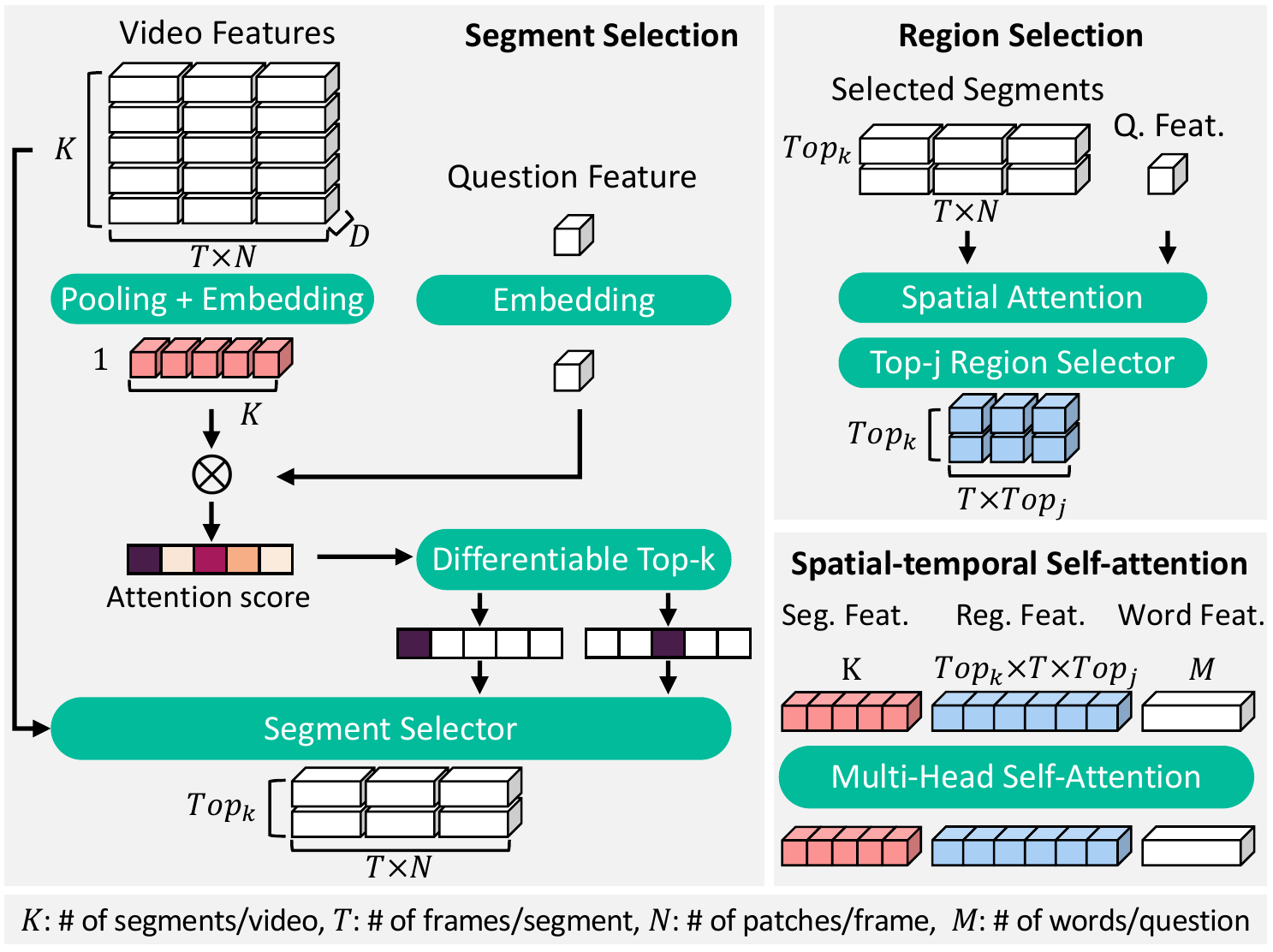}
\vspace{-0.7cm}
\caption{\textbf{Key components of Iterative Spatial-Temporal Attention Layer.} Since region selection follows the same architecture as segment selection, we only show its inputs and outputs.}
\label{fig:ISTA}
\vspace{-0.3cm}
\end{figure}

The Iterative Spatial-Temporal Attention layer (ISTA) aims to iteratively select the segments and regions among a long video conditioned on questions and then perform multi-event reasoning over selected ones. Specifically, ISTA contains three steps: segment selection, region selection, and spatial-temporal self-attention, as shown in Fig.~\ref{fig:ISTA}.

\textbf{Segment Selection.}
Given a set of image patch features $\boldsymbol{x}$, we calculate the features of segments and the question, then select the patch features of $Top_k$ segments by performing cross-modal temporal attention and differentiable top-k selection. 

Specifically, to perform temporal attention, the frame features are first obtained by pooling the features in spatial dimension: the $t$-th frame feature in $k$-th segment is calculated as $f_t^k = \mathrm{pool}(x_{t, 1}^k, x_{t, 2}^k, ..., x_{t, N}^k)$, where $x_{t, n}^k$ indicates $n$-th patch at $t$-th frame of $k$-th segment. Then, the segment features are obtained by pooling frames features along the temporal dimension: $s^k = \mathrm{pool}(f_{1}^k, f_{2}^k, ..., f_{T}^k)$. The question feature is similarly obtained by pooling the word features, $\boldsymbol{q} = \mathrm{pool}(w_1, ..., w_M)$. The pooling functions can be chosen from mean pooling, first token pooling, simple MLP layer, etc., according to the specific type of used vision-language Transformer. For example, for image-language Transformers, like CLIP, the first token pooling can be used for extracting frame and question features and mean pooling over frames for obtaining segment features. 

Given the segment features $\mathbf{S} = \{s^k\}_{k=1}^K$, patch features $\mathbf{X} = \{x^k\}_{k=1}^K$, and question features $\boldsymbol{q}$, we first perform cross-modal temporal attention among $\mathbf{S}$ given $\mathbf{q}$, and then conduct top-k feature selection over $\mathbf{X}$, as formulated:
\begin{align}
    &\mathbf{Q} = g_q(\boldsymbol{q}),  
    \mathbf{K} = g_s(\mathbf{S}),  
    \mathbf{V} = \mathbf{X}, \\
    &\mathbf{X}_t = \underset{Top_k}{\mathrm{selector}}(\mathrm{softmax}(\frac{\mathbf{QK}^T}{\sqrt{d_k}}), \mathbf{V}),
\end{align}
where $g_q$ and $g_s$ are linear projection layers for different types of features, $\mathrm{selector}$ is a differentiable top-k selection function to choose the spatial features of $Top_k$ segments. 
The top-k selection can be implemented by expanding the Gumbel-Softmax trick~\cite{jang2016categorical} or based on optimal-transport formulations ~\cite{xie2020differentiable} for ranking and sorting. In this paper, we simply conduct Gumbel-Softmax sampling $Top_k$ times with replacement to achieve top-k selection. Note that we sample the segments with replacement because, in some cases, the question could only involve one segment. We hope the model learns to enhance the most related segment in such cases by re-sampling it, instead of forcing it to select an irrelevant segment, as sampling without replacement will do. See \supplement for more discussion about Top-k selection.
The output of the module is $\mathbf{X}_t = \{x^k | k \in \mathcal{B}\} \in \mathbb{R}^{Top_k \times T \times N \times D}$, where $\mathcal{B}$ is the set of selected $Top_k$ segments' indexes.

\textbf{Region Selection.}
For the $\tau$-th sampled frame, we want to select its most relevant patches with the question. Given its region feature of one frame $\mathbf{X}_{\tau} = \{x_{\tau, n}^k| n \in [1, N], k\in \mathcal{B}\}$ along with question $\mathbf{q}$, we perform cross-model attention over spatial patches of the $\tau$-th sampled frame and select the $Top_j$ most related patches. This can be formulated as:
\begin{align}
    &\mathbf{Q} = h_q(q),  
    \mathbf{K} = h_{x}(\mathbf{X}_{\tau}),  
    \mathbf{V} = \mathbf{X}_{\tau}, \\
    &\mathbf{X}_{\tau}' = \underset{Top_j}{\mathrm{selector}}(\mathrm{softmax}(\frac{\mathbf{QK}^T}{\sqrt{d_k}}), \mathbf{V}),
\end{align}
where $h_q$ and $h_{x}$ are embedding layers for linear feature projection. The output of the given each frame is $\mathbf{X}_{\tau}' \in \mathbb{R}^{Top_j \times D}$. Finally, we stack the selected patch features of all selected frames to obtain $\mathbf{X}_{st} = \{\mathbf{X}_{\tau}' | \tau \in [1, Top_k \times T]\}$.

\textbf{Spatial-Temporal Self-Attention.}
Given the selected frames and selected regions, along with the question, we aim to employ a self-attention layer to reason out a fused feature vector to jointly represent the question and video.

Regards the inputs of self-attention, since the main computation cost comes from too many patches ($K \times T \times N$, about thousands of patches), we only keep the selected ones. While for temporal information, we keep all segments as the total number is only $K$ (usually less than 10), which doesn't bring heavy cost and can benefit more comprehensive multi-event reasoning.

Specifically, we first add type embedding to indicate the types feature, e.g., image region, segment, or word. The type embedding is formulated $P_h \in \{\phi_h(h) | h \in [1, 3]\}$ to each feature for indicating  where $\phi_h$ is a trainable embedding layer. Then, a standard multi-head attention is performed to obtain the contextual features of all input tokens, formulated as:
\begin{align}
  \mathbf{X}^{(i)}_{o} = \mathrm{MultiHead}([\phi_s(\mathbf{S}); \phi_x(\mathbf{X}_{st}); \phi_w(\mathbf{X}_w)]),
\end{align}
where $\phi_s$, $\phi_x$, and $\phi_w$ are linear transformation. 

\textbf{Iterative Execution of ISTA.} A stack of $L$ ISTA layers is used for modelling multi-event interactions between a given question and video, where the updated segment features and word features are fed into the next layer. The output of each layer $\{\mathbf{X}_o^{(l)}\}_{l=1}^L$ is used for answer prediction.

\subsection{Answer Prediction}
\label{sec3.3}

Specifically, we mean pool the token features of all ISTA layers, $\mathbf{X}_o = \mathrm{MeanPool}(\mathbf{X}_o^{(1)}, ..., \mathbf{X}_o^{(L)})$. In addition, following the work~\cite{yang2021just}, we calculate the similarity between the $\mathbf{X}_o$ and the feature of all candidate answers $\mathbf{X}_A=\{\mathbf{x}_a|a\in \mathcal{A}\}$ obtained by using the pre-trained model. Finally, the candidate answer with the maximal similarity is considered as the final prediction $\widetilde{y}$.
\begin{align}
    \widetilde{y} = { \underset{y\in \mathcal{A}}{\arg \max}}(\mathbf{X}_o (\mathbf{X}_A)^T).
\end{align}
During training, we optimize the softmax cross-entropy loss between the predicted similarity scores and ground truth.

%% file: section/exp.tex
\section{Experiments}
\subsection{Datasets}
We evaluate our model on four recently proposed challenging datasets for the long-form VideoQA, namely AGQA~\cite{grunde2021agqa}, NExT-QA~\cite{xiao2021next}, STAR~\cite{wu2021star} and Env-QA~\cite{gao2021env}. 

\textbf{AGQA} is an open-ended VideoQA benchmark for compositional spatio-temporal reasoning. We use its v2 version, which has more balanced distributions, as the dataset creator recommended. It provides 2.27M QA pairs over 9.7K videos with an average length of 30 seconds.

\textbf{NExT-QA} is a multi-choice VideoQA benchmark for causal and temporal reasoning. It contains a total of 5,440 videos with an average length of 44s and about 52K questions.

\textbf{STAR} is another multi-choice VideoQA benchmark for Situated Reasoning. STAR contains 22K video clips with an average length of 12s, along with 60K questions.

\textbf{Env-QA} is an open-ended VideoQA benchmark for dynamic environment understanding. It contains 23K egocentric videos with an average length of 20 seconds collected on virtual environment AI2THOR~\cite{kolve2017ai2} along with 85K questions.

\begin{table*}[]
\centering
\resizebox{0.9\textwidth}{!}{%
\begin{tabular}{l|cccccc|cc}
\toprule
Question Types       & Most Likely & PSAC  & HME   & HCRN~\cite{le2020hierarchical}      &AIO~\cite{wang2022all} & Temp[ATP]~\cite{buch2022revisiting} & \modelname- AIO & \modelname- CLIP \\ \midrule
Object-relation          & 9.39        & 37.84 & 37.42 & 40.33 & 48.34  & 50.15     & 51.43                         &    \textbf{51.68}        \\
Relation-action            & 50.00       & 49.95 & 49.90 & 49.86 & 48.99  & 49.76     & 54.67                         &    \textbf{67.18}        \\
Object-action            & 50.00       & 50.00 & 49.97 & 49.85 & 49.66  & 46.25     & 55.37                         &    \textbf{68.99}        \\
Superlative          & 21.01       & 33.20 & 33.21 & 33.55 & 37.53  & 39.78     & 41.34                        &    \textbf{42.05}         \\
Sequencing          & 49.78       & 49.78 & 49.77 & 49.70 & 49.61  & 48.25     & 53.14                         &    \textbf{67.24}          \\
Exists               & 50.00       & 49.94 & 49.96 & 50.01 & 50.81  & 51.79     & 53.49                         &    \textbf{60.33}        \\
Duration comparison  & 24.27       & 45.21 & 47.03 & 43.84 & 45.36  & 49.59     & 47.48                         &    \textbf{54.62}        \\
Activity recognition & 5.52        & 4.14  & 5.43  & 5.52  & 18.97  & 18.96     & \textbf{20.18}                         &    19.69        \\
\midrule
All                        & 10.99       & 40.18 & 39.89 & 42.11 & 48.59  & 49.79     & 50.96                         &    \textbf{54.39}              \\ 
\bottomrule
\end{tabular}%
}
\vspace{-0.3cm}
\caption{QA accuracies of state-of-the-art (SOTA) methods on AGQA v2 test set.}
\vspace{-0.3cm}
\label{tab:agqa}
\end{table*}

\begin{table}[]
\centering
\resizebox{0.45\textwidth}{!}{%
\begin{tabular}{l|ccc|c}
\toprule
\multicolumn{1}{c|}{Method}      & Causal & Temporal & Descriptive & All   \\ \midrule
HGA                                                           & 44.22  & 52.49    & 44.07       & 49.74 \\
CLIP (single frame)                                           & 46.3   & 39.0     & 53.1        & 43.7  \\
VQA-T~\cite{yang2021just}               & 49.60  & 51.49    & 63.19       & 52.32 \\
AIO~\cite{wang2022all}                  & 48.04  & 48.63    & 63.24       & 50.60 \\
Temp[ATP]~\cite{buch2022revisiting}     & 48.6   & 49.3     & 65.0        & 51.5  \\
Temp[ATP]+ATP~\cite{buch2022revisiting} & 53.1   & 50.2     & 66.8        & 54.3  \\
VGT~\cite{xiao2022video}                & 52.28  & 55.09    & 64.09       & 55.02 \\
\midrule
\modelname - AIO                        & 51.54  & 51.63    & 64.16       & 53.54 \\
\modelname - CLIP                       & \textbf{54.62}  & \textbf{56.64}  & \textbf{66.92}  & \textbf{57.18} \\
\bottomrule
\end{tabular}%
}
\vspace{-0.3cm}
\caption{QA accuracies of SOTA methods on NExT-QA val set.}
\vspace{-0.3cm}
\label{tab:nextqa}
\end{table}

\begin{table}[th]
\resizebox{0.475\textwidth}{!}{%
\begin{tabular}{l|cccc|c}
\toprule
\multicolumn{1}{c|}{Method}           & Interaction & Sequence & Prediction & Feasibility & Mean \\
\midrule
ClipBERT~\cite{lei2021less}           & 39.81        & 43.59     & 32.34       & 31.42        & 36.7    \\
CLIP~\cite{radford2021learning}       & 39.8        & 40.5     & 35.5       & 36.0        & 38.0    \\
RESERVE-B~\cite{zellers2022merlot}    & 44.8        & 42.4     & 38.8       & 36.2        & 40.5    \\
Flamingo-9B~\cite{alayrac2022flamingo}    & -       & -        & -          & -           & 43.4    \\
AIO~\cite{wang2022all}                & 47.53       & 50.81    & 47.75      & 44.08       & 47.54 \\ 
Temp[ATP]~\cite{buch2022revisiting}   & 50.63       & 52.87    & 49.36      & 40.61       & 48.37  \\ \midrule
\modelname - AIO                      & 53.00       & 52.37    & 49.52      & 43.87       & 49.69 \\ 
\modelname - CLIP                     & \textbf{55.59}  & \textbf{54.23}    & \textbf{54.24}      & \textbf{44.48}       & \textbf{51.13}    \\
\bottomrule
\end{tabular}%
}
\vspace{-0.3cm}
\caption{QA accuracies of SOTA methods on STAR val set.}
\vspace{-0.3cm}
\label{tab:star}
\end{table}

For each benchmark, we follow standard protocols outlined by prior works~\cite{grunde2021agqa, buch2022revisiting, alayrac2022flamingo, gao2021env} for dataset processing, metrics, and settings. Please see \supplement for details.

\subsection{Implementation Details}
Our proposed method can be built upon most of the pre-trained multi-modal Transformers. In our experiments, we try two typical types of pre-trained models, CLIP (ViT-B/32)~\cite{radford2021learning} for image-language pre-training models and All-in-One-Base~\cite{wang2022all} for video-language pre-training model, denoted as \modelname-CLIP and \modelname-AIO respectively. In\modelname, we set $Top_k=2$ and $Top_j=12$ in cascade selection module and the layer of ISTA $L=2$. For all videos, we sample 32 frames per video, and split them into $K=8$ segments. AdamW is utilized to optimize model training. Our model is trained on NVIDIA RTX A5000 GPUs and implemented in PyTorch.\footnote{The code will be available at \url{https://github.com/showlab/mist}.}

\begin{table}[ht]
\centering
\resizebox{0.475\textwidth}{!}{%
\begin{tabular}{l|ccccc|c}
\toprule
\multicolumn{1}{c|}{Method}                       & Attribute & State & Event & Order & Number & All \\ \midrule 
CNN-LSTM                   & 38.21     & 42.26 & 29.94 & 53.37 &  38.12 & 38.05   \\
ST-VQA~\cite{jang2019video}& 41.66 & 48.98 & 33.87 & 54.09 &  38.54 & 41.97   \\
STAGE~\cite{lei2020tvqa+}  & 39.49 & 49.93 & 34.52 & 55.32 &  37.98 & 42.53   \\ 
AIO~\cite{wang2022all}               & 41.78 & 52.98 & 37.57 & 55.16  & 38.50  & 44.86  \\ 
Temp[ATP]~\cite{buch2022revisiting}   & 42.87  & 53.49   & 38.35  & 55.25  & 38.65  & 45.43  \\ 
TSEA~\cite{gao2021env}               & 42.96 & 56.73 & 39.84 & 55.53 &  39.35 & 47.06  \\
\midrule
\modelname-AIO                 &  43.63     & 55.17      & 40.99      & 55.44      &  39.54      & 47.19 \\ 
\modelname-CLIP                & \textbf{44.05} & \textbf{58.13}  & \textbf{42.54}  & \textbf{56.83}   & \textbf{40.32}        & \textbf{48.97} \\
\bottomrule
\end{tabular}%
}
\vspace{-0.3cm}
\caption{QA accuracies of SOTA methods on Env-QA test set.}
\label{tab:envqa}
\vspace{-0.3cm}
\end{table}

\subsection{Comparison with State-of-the-arts}
We compare our model with the state-of-the-art (SOTA) methods on four VideoQA datasets (i.e., AGQA v2, NExT, STAR, and Env-QA), as shown in Tab.~\ref{tab:agqa}, ~\ref{tab:nextqa}, ~\ref{tab:star}, and \ref{tab:envqa} respectively. We can see that our proposed method achieves state-of-the-art performances and outperforms the existing methods on all datasets. The performance gain is relatively limited on Env-QA, because its videos are recorded in a virtual environment, AI2THOR. There is a domain gap for CLIP feature, while previous SOTA uses the features pre-trained on virtual environment data. 

Notably, among SOTAs, TEMP[ATP]~\cite{buch2022revisiting} uses the same feature, CLIP (ViT-B/32), as \modelname-CLIP. And All-in-one~\cite{wang2022all} and \modelname-AIO also use the same feature, All-in-One-Base.
Compared to these methods, it can be found that our two versions of models, which build upon different types of pre-trained models, achieve substantial performance gains on all datasets.

Moreover, from the question type breakdown of each dataset, we can see that compared with AIO and Temp[ATP], our model obtains a much more significant performance boost on questions that require multi-grained visual concepts reasoning (i.e., \emph{Rel.-act.}, \emph{Obj.-act.} on AGQA v2) than those which mainly require information within one frame (i.e., \emph{Obj.-rel.} on AGQA v2 and \emph{Descriptive} on NExT-QA). 
In addition, we can see that our model surpasses these models with large margin on questions requiring causality or multi-event reasoning, e.g., \emph{Sequencing} in AGQA v2, \emph{Causal} \& \emph{Temporal} in NExT-QA, \emph{Interaction} \& \emph{Prediction} in STAR, and \emph{Event} in Env-QA. These results demonstrate that our proposed model can effectively address the unique challenges of long-form video QA.

\begin{table}[]
\centering
\resizebox{0.475\textwidth}{!}{%
\begin{tabular}{l|ccc|cccc}
\toprule
\multicolumn{1}{c|}{\multirow{2}{*}{Method}} & \multicolumn{3}{c|}{AGQA v2}                    & \multicolumn{4}{c}{NExT-QA}   \\ 
\multicolumn{1}{c|}{}                        & Binary & Open  & \multicolumn{1}{c|}{All}   & C.    & T.    & D.    & All   \\ \midrule
MeanPool                                     & 49.26  & 34.01 & \multicolumn{1}{c|}{41.58} & 47.87 & 45.22 & 58.01 & 48.59 \\
Trans.-Frame                                 & 54.03  & 45.66 & \multicolumn{1}{c|}{49.66} & 50.77 & 49.96 & 65.27 & 52.76 \\
Trans.-Patch                                 & 55.09  & 47.08 & \multicolumn{1}{c|}{51.05} & 52.58 & 50.42 & 64.55 & 53.74 \\
Divided STA                                  & 55.93  & 46.88 & \multicolumn{1}{c|}{51.37} &  52.03     & 50.24     & 64.31    & 53.36    \\ \midrule
\modelname- CLIP   & 58.28 & 50.56     &  54.39 & 54.62  & 56.64  & 66.92  & 57.18 \\ \bottomrule
\end{tabular}%
}
\vspace{-0.3cm}
\caption{QA accuracies of variants of \modelname on AGQA v2 and NExT-QA.}
\label{tab:mist-ablate}
\vspace{-0.3cm}
\end{table}

\subsection{Comparison with Baselines}
Here we devise several alternative solutions for long-form video modelling to replace our proposed ISTA.
Specifically, in our CLIP-based \modelname framework, we compare ISTA against other solutions, by fine-tuning the same pre-training input representation on AGQA v2 dataset.
\begin{itemize}[itemsep=2pt,topsep=2pt,parsep=0pt,leftmargin=12pt]
\item MeanPool: It simply takes the average of frame features as the representation of the whole video.
\item Trans.-Frame: We follow the seqTransf type in CLIP4Clip, utilizing a Transformer to perform self-attention over frames features to represent the video.
\item Trans.-Patch: This model is similar to Trans.-Frame, but it perform self-attention over all patch tokens.
\item Divided STA: We follow TimeSformer~\cite{bertasius2021space} in video classification model to perform uni-modal two-step Space-Time Attention over image patches.
\end{itemize}

From the results in Tab.~\ref{tab:mist-ablate}, we can see that ISTA achieves substantial improvement over other variants with larger than 3\% improvement on the overall accuracy. In addition, we find that for long-form VideoQA, the Transformer-based answer prediction models are much better than the MeanPool method, while in the video-text retrieval field, sometimes mean pooling is even better. The reason could be that the content of a long-form video is often complex and diverse, and a simple method for aggregating all frame features, such as mean pooling, may cause information loss. And long-form video QA requires more powerful temporal and spatial reasoning ability to focus on some details of a video, while mean pooling only performs well on capturing overall content. 

Moreover, we can see that it is helpful to consider region information in long-form QA (Divided STA and Trans.-Path outperform Trans.-Frame). But, neither dense self-attention nor divided STA considers the interaction among multi-grained concepts; thus, the performance improvement is limited. And after integrating different granularities of visual concepts during reasoning, our method benefits the performance.
All the above findings show that our method is effective, and transferring pre-trained transformers to long-form video QA is a challenging topic worth exploring.

\subsection{Ablation Study}
In this section, we propose several sets of variants of \modelname to show the effectiveness of its key components.

\begin{table}[]
\centering
\resizebox{0.475\textwidth}{!}{%
\begin{tabular}{l|ccc|cccc}
\toprule
\multicolumn{1}{c|}{\multirow{2}{*}{Method}} & \multicolumn{3}{c|}{AGQA v2}                    & \multicolumn{4}{c}{NExT-QA}   \\  
\multicolumn{1}{c|}{}                        & Binary & Open  & \multicolumn{1}{c|}{All}   & C.    & T.    & D.    & All   \\ \midrule
\modelname w/o. SS             & 55.37 & 47.50 & 51.40  & 51.24     & 51.39     & 65.43    & 53.49                \\
\modelname w/o. RS             & 58.18 & 50.14 & 54.13 & 54.32     & 56.14     & 66.56    & 56.81              \\
\modelname w/o. STA        & 50.93 & 36.75 & 43.79  & 48.99     & 43.92     & 60.37    & 49.12               \\\midrule
\modelname- CLIP   & 58.28 & 50.56     &  54.39 & 54.62  & 56.64  & 66.92  & 57.18 \\ \bottomrule
\end{tabular}%
}
\vspace{-0.3cm}
\caption{Ablations results of ISTA on AGQA v2 and NExT-QA.}
\label{tab:ista-ablate}
\vspace{-0.3cm}
\end{table}

\textbf{Effect of each component in ISTA.}
We ablate key modules in ISTA layer, i.e., Segment Selection, Region Selection, or Self-attention layer, denoted as, \modelname w/o. SS/RS/STA, respectively: 
\begin{itemize}[itemsep=2pt,topsep=0pt,parsep=0pt,leftmargin=12pt]
\item \modelname w/o. SS: It removes the Segment Selection module, and only performs region selection. Patch features with word features are fed into self-attention module.
\item \modelname w/o. RS: It removes Segment Selection module. All region features within selected segments are fed into self-attention layer.
\item \modelname w/o. STA: The segment features and selected region features are mean pooled as the output of ISTA.
\end{itemize}

\begin{figure}[t]
\centering
\includegraphics[width=0.45\textwidth]{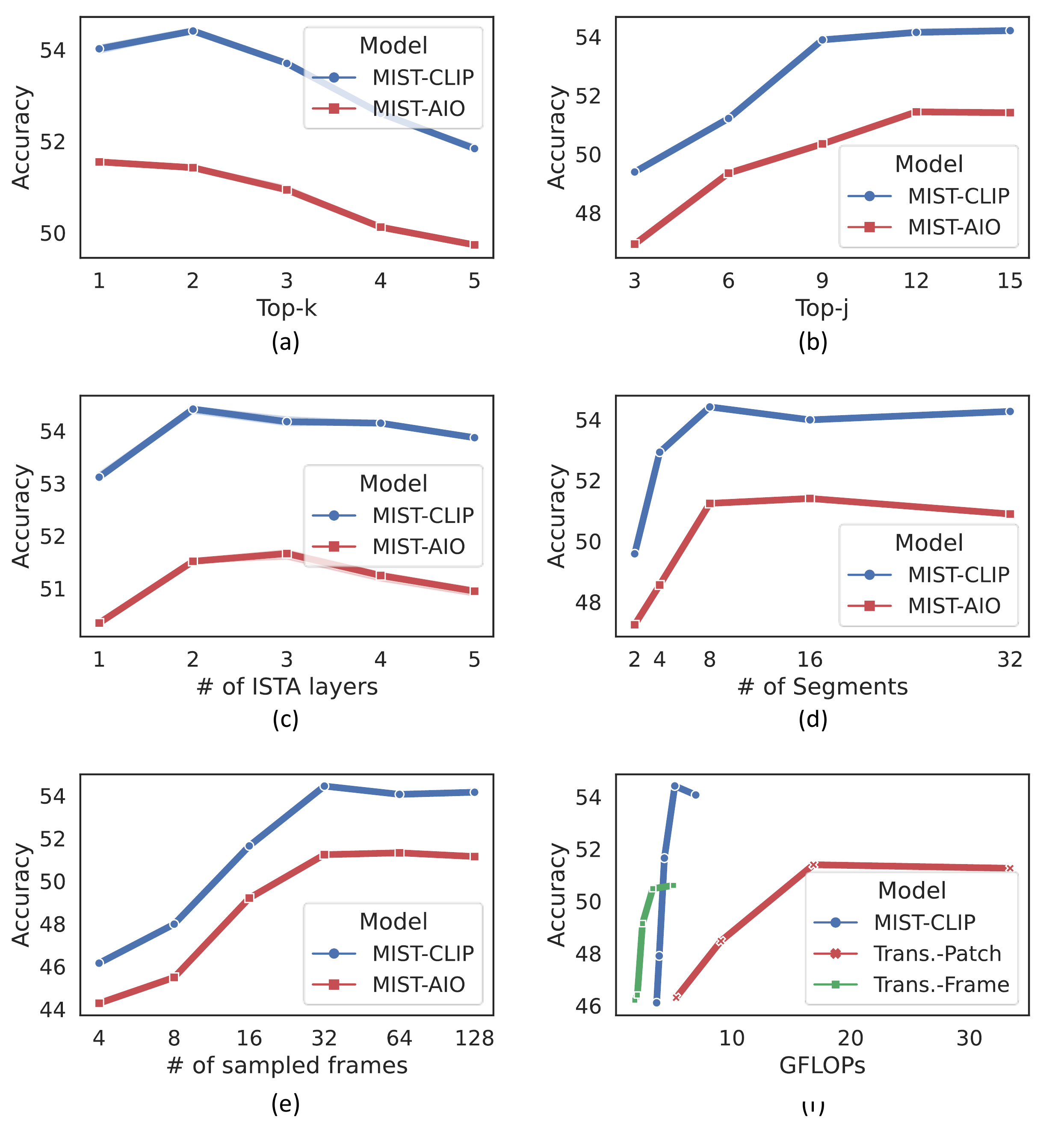}
\vspace{-0.4cm}
\caption{\textbf{Performances of \modelname with different settings.} (a-e) Performances of \modelname with different hyper-parameters on AGQA v2. (f) Performance of variants of\modelname under different GFLOPs on AGQA v2, where GFLOPs rise with the number of sampled frames increase.}
\vspace{-0.4cm}
\label{fig.config}

\end{figure}

\begin{figure*}[t]
\centering
\includegraphics[width=0.95\textwidth]{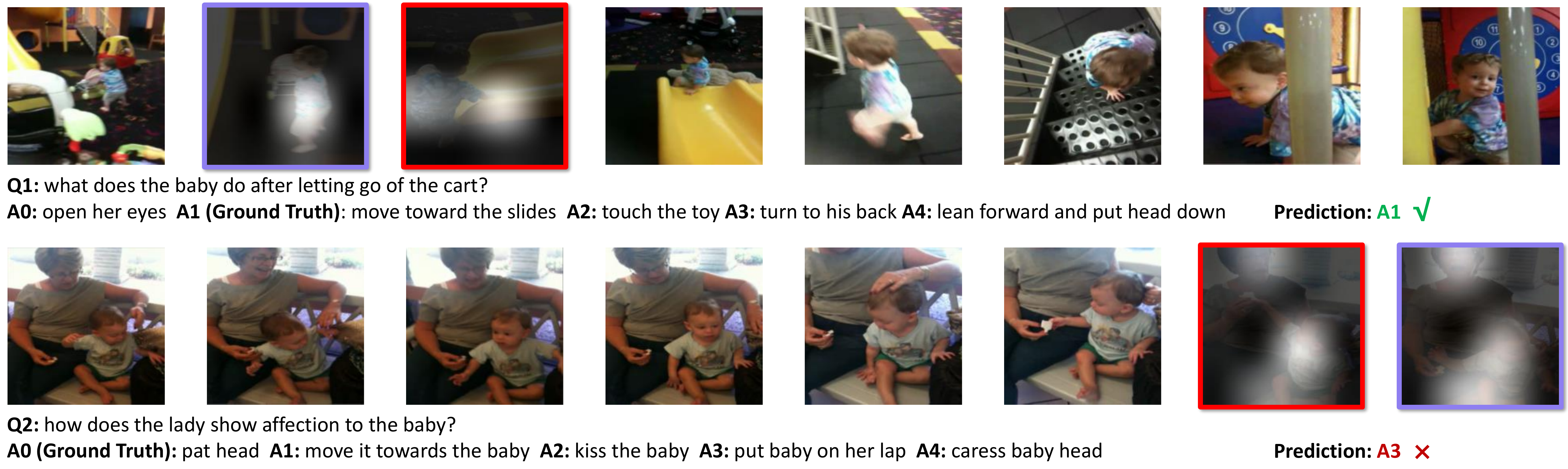}
\vspace{-0.3cm}
\caption{\textbf{Qualitative results of \modelname on NExT-QA dataset.} We visualize its prediction results along with spatial-temporal attention, where the frames with purple and red outlines indicate the highest temporal attention score in the first and second ISTA layers, respectively.}
\vspace{-0.5cm}
\label{fig.example}
\end{figure*}

The results of these variants on AGQA v2 and NExT-QA are shown in Tab.~\ref{tab:ista-ablate}. We can see that removing Segment Selection causes a larger than 3\% accuracy drop. The reason could be that removing it will introduce a lot of irrelevant region information when predicting the answer and thus hurt the performance. Tab.~\ref{tab:ista-ablate} also shows that Segment Selection is important for multi-event reasoning because removing it hurts the performances on questions requiring temporal reasoning, i.e., \emph{Causal} and \emph{Temporal}.

In addition, the performance drop on both datasets is significant when removing Spatial-temporal self-attention. The reason may be similar to MeanPool. We need a powerful model to capture multi-grained reasoning.

Moreover, we can see that removing spatial attention doesn't hurt performance too much. The number of objects in the video frames is relatively small (compared with natural scene images in image QA), and after temporal attention, the patch number has already been greatly reduced. So, the existing model is able to effectively focus on the appropriate objects. But, It is worth mentioning that we can reduce the computation cost by using a spatial selection module. It may be useful when we face high-resolution or extremely complex videos in the future.

\textbf{Effects of different ISTA configurations.}
In this part, we try different configurations of model architecture, including number of selected segments $Top_k$, select patches $Top_j$, ISTA layers $L$, and the number of segments $K$. The results are show in Fig.~\ref{fig.config} (a-d). 

First, Fig.~\ref{fig.config} (a) shows that the performance is relatively good under the small $Top_k$. 1 or 2 are good choices on two datasets. The performance slightly drops if $Top_k$ further increases. It indicates that considering too many segments in reasoning will introduce redundant information, causing performance drops. For the number of selected patches $Top_j$, as shown in Fig.~\ref{fig.config} (b), we can see that with the increase of $Top_j$, the performance first increases, then reaches stability. The reason for this phenomenon could be that when selecting too few image regions, it may incorrectly filter some regions used for answering questions. And when the selected regions increase, though it introduces some irrelevant regions, since the patch number after segment selection are already relatively small, the self-attention module can effectively attend to relevant regions. 

For the number of ISTA layers, as shown in Fig.~\ref{fig.config} (c), with the increase of $L$, the performance increases first and then reaches stability or slightly drops. It shows that stacking several layers of ISTA can benefit multi-event reasoning. In addition, the performance doesn't constantly increase with larger $L$. This is probably due to (1) the datasets are not large enough to train a deeper network and (2) the questions usually only involve two or three events, so considering more events may not bring more benefits. Fig.~\ref{fig.config} (d) shows that when varying the number of video segments, performance tends to suffer when the videos are under-segmentation, because in this case, each segment spans a relatively long duration, and hence the Segment Selection module is useless. More importantly, all those findings imply that\modelname is effective in multi-event reasoning by attending to multiple segments.

\subsection{Computation Efficiency}
In Fig.~\ref{fig.config} (e), we can see that the accuracy increases significantly when sampling more frames. It indicates that sampling more frames for long video QA tasks could be necessary. Though current datasets don't provide videos with several minutes or hours duration, such long videos are likely to be encountered in real application scenarios. Efficiency issues thus could be a more crucial consideration in such cases. In Fig.~\ref{fig.config} (f), we compare GFLOPs vs. accuracy for ours against other long-form video QA methods. It can be seen that the standard Transformer over patches is computationally expensive. The frame-based method is lightweight in computation, but its performance is limited. Our method requires only a little extra computation but achieves much better performance. It is also worth mentioning that\modelname doesn't enlarge model size for higher efficiency. Compared with other methods, it only contains some extra shallow networks for spatial-temporal attention.

\subsection{Qualitative Results}
We visualize some success and failure cases from the NExT-QA dataset in Fig.~\ref{fig.example}. It can be seen that our model can explicitly select video clips and image regions relevant to the question. We can also find that it is difficult for the model to correctly select segments and regions, when the question mainly involves some concepts related to social emotions. Existing pre-trained models may not well understand the correspondence between abstract concepts and videos. However, we believe that these issues can be alleviated by proposing better pre-trained models on short videos, and our method is easy to build upon the stronger ones.

%% file: section/conclusion.tex
\section{Conclusion and Future Work}
This paper introduces\textcinzelit{M}ulti-modal\textcinzelit{I}terative\textcinzelit{S}patial-temporal\textcinzelit{T}ransformer for long-form VideoQA, which decomposes dense self-attention into a cascade segment and region selection module to increase the computation efficiency along with a self-attention layer to reason over various grained visual concepts. In addition, by iteratively conducting selection and attention over layers, \modelname better performs multi-event reasoning. Experimental results on four VideoQA datasets show its effectiveness and advantages in efficiency and interpretability. For future work, although \modelname has increased the number of sample frames, the ability to capture high-frequency motion may still need to be improved. In addition, patch features naturally have some limitations in complex object-level reasoning. Recently, there have been some pre-trained models for specifically modeling actions and objects. It may be interesting to try more types of pre-trained models or even combine many of them to achieve more general reasoning.